\documentclass[letterpaper]{article} 
\usepackage{emnlp2021}
\usepackage{times}  
\usepackage{helvet}  
\usepackage{courier}  

\usepackage{algorithm}
\usepackage{algpseudocode}

\usepackage{cleveref}

\usepackage{subcaption}
\usepackage{pgfplots}
\usepackage[utf8]{inputenc}
\usepackage{pgfplots}
\DeclareUnicodeCharacter{2212}{−}
\usepgfplotslibrary{groupplots,dateplot}
\usetikzlibrary{patterns,shapes.arrows}

\newcommand{\gdirect}{$\overline{G}_{{\mbox{direct}}}$}
\newcommand{\gpartial}{$\overline{G}_{{\mbox{partial}}}$}
\newcommand{\gscratch}{$\overline{G}_{{\mbox{scratch}}}$}
\pdfoutput=1

\usepackage{booktabs, multirow} 
\usepackage{soul}
\usepackage{changepage,threeparttable} 

\usepackage{newfloat}
\usepackage{listings}
\floatstyle{ruled}
\newfloat{listing}{tb}{lst}{}
\floatname{listing}{Listing}

\pgfplotsset{compat=1.17}
\begin{document}
%
\title{SuperShaper: Task-Agnostic Super Pre-training \\ of BERT Models with Variable Hidden Dimensions}
\author{
    Vinod Ganesan$^{1,3*}$ \ Gowtham Ramesh$^{2}$\thanks{\ \ The first two authors have contributed equally.} \  Pratyush Kumar$^{1,3}$
    \\ \\
     $^1$Indian Institute of Technology, Madras
    $^2$Robert Bosch Center for Data Science and Artificial Intelligence, \\
    $^3$Microsoft Research, India \\
    vinodg@cse.iitm.ac.in, gowtham.ramesh1@gmail.com, kumar.pratyush@microsoft.com
   
}
 
\maketitle

\begin{abstract}
\begin{quote}
\noindent 
Task-agnostic pre-training followed by task-specific fine-tuning is a default approach to train NLU models. 
Such models need to be deployed on devices across the cloud and the edge with varying resource and accuracy constraints. 
For a given task, repeating pre-training and fine-tuning across tens of devices is prohibitively expensive. 
We propose SuperShaper, a task agnostic pre-training approach which simultaneously pre-trains a large number of Transformer models by varying shapes, i.e., by varying the hidden dimensions across layers. 
This is enabled by a backbone network with linear bottleneck matrices around each Transformer layer which are sliced to generate differently shaped sub-networks.
In spite of its simple design space and efficient implementation, SuperShaper discovers networks that effectively trade-off accuracy and model size: Discovered networks are more accurate than a range of hand-crafted and automatically searched networks on GLUE benchmarks.
Further, we find two critical advantages of shape as a design variable for Neural Architecture Search (NAS): (a) heuristics of \textit{good shapes} can be derived and networks found with these heuristics match and even improve on carefully searched networks across a range of parameter counts, and (b) the latency of networks across multiple CPUs and GPUs are insensitive to the shape and thus enable device-agnostic search.
In summary, SuperShaper radically simplifies NAS for language models and discovers networks that generalize across tasks, parameter constraints, and devices.
\end{quote}
\end{abstract}

\section{Introduction}
\label{sec:intro}
Computing over the last 60 years has swung multiple times between increased centralization and increased distribution.
The last decade, with a surge in public and private cloud usage, has centralized compute and storage.
But rising cloud costs, ever powerful client devices, and increased call for privacy are swinging the pendulum back to more compute on the edge.
This ongoing transition is specifically relevant for compute-intensive AI inference.
Caught between both ends of the swinging pendulum, AI model owners must ensure that their models are deployable on tens of different devices spanning CPU and GPU setups on cloud and client devices.

This wide deployment surface is particularly challenging for NLP applications for two reasons. 
First, NLU models, for instance for sentiment classification, are trained with a hybrid pipeline that combines pre-training and finetuning, where the former is significantly more expensive than the latter.
For instance, pre-training BERT on C4 takes 1-5 days on 128 TPUv2s \cite{Liu2021PayAT}.
In comparison, end-to-end training of GMLP (with similar model parameters) on ImageNet takes 4 hours \cite{Liu2021PayAT}.
Second, models are usually deployed for multiple NLU tasks such as sentiment classification, paraphrase detection, question answering, natural language inference, etc.
For every point in the product space of multiple tasks and multiple devices, different model variants have to be pretrained and evaluated, which would be very expensive.

There are three broad solutions to address this optimization complexity.
First, a single large language model, such as BERT, could be trained agnostic of task and device.
Then for every device and task under consideration, the model can be sized down with techniques such as knowledge distillation \cite{Tang2019DistillingTK, Turc2019WellReadSL, Sanh2019DistilBERTAD, jiao2020tinybert}, pruning \cite{michel2019sixteen,goyal2020power}, quantization \cite{shen2020q}, factorization \cite{ma2019tensorized}, etc.
This approach does not generalize across devices or tasks and retains a large search complexity. 
Also, model compression is inherently lossy in not retaining all information learnt during pre-training. 
The second approach is to pretrain a single language model but simultaneously finetune many sub-networks of different sizes, in what is called \textit{super fine-tuning}. 
Then for a chosen task and device an appropriately sized sub-network can be sampled from the super-network and deployed.
The third approach takes this even further: Instead of pre-training one large language model, an entire family of language models is trained, in what is called \textit{super pre-training}.

Over the last year, two works have studied task-specific super fine-tuning, namely DynaBERT \cite{ludyna2020} and YOCO-BERT \cite{zhang2021you}, while task-agnostic super pre-training has been studied in NAS-BERT \cite{xu2021bert}.
In all three cases, starting from a BERT model, different models are trained as defined by an architectural search space.
For instance, in DynaBERT, the number of attention heads and Transformer layers are changed, in NAS-BERT multi-headed attention, separable convolution, and feed-forward networks are independently scaled and combined, and in YOCO-BERT the number of layers, size of representations in FFN, and number of attention heads are changed.
Given these large and multi-dimensional search spaces, each method proposes complex algorithms to super-train.
DynaBERT proposes knowledge distillation and an importance based rewiring across layers to retain accurate sub-networks.
NAS-BERT proposes reducing the search space by discretizing the network into blocks, and a heuristic based search space pruning.
YOCO-BERT proposes an iterative explore-exploit algorithm that maintains a dynamic set of promising sub-networks which are trained simultaneously.

We propose an alternative approach to super pre-training language models, called SuperShaper.
Like NAS-BERT, SuperShaper is task-agnostic and trains a family of language models.
However, SuperShaper differs from existing methods in two crucial ways.
First, the search space for SuperShaper is much simpler and includes only the hidden dimension of each Transformer layer, which we refer to as the \textit{shape} of the network.
This is enabled by modifying the BERT backbone with bottleneck matrices at the input and output of each layer, inspired from the efficient model MobileBERT \cite{sun-etal-2020-mobilebert}.
In each batch, differently shaped networks are randomly sampled by slicing the bottleneck matrices and trained.
Though a single parameter per layer, the hidden dimension sensitively affects model capacity as the parameter count linearly depends on it.
Second, the super pre-training procedure is much simpler with SuperShaper and requires only sliced matrix multiplications on the bottleneck matrices, similar to the earliest techniques proposed for elastic training \cite{brock2018smash,cai2019once}.
This is radically simpler than existing NAS techniques which define complex design spaces, architecture modifications, and heuristics for managing the search space.
Indeed, in a PyTorch code-base only about 20 lines of additional code are required to add SuperShaper functionality. 

Despite the simple design space and efficient implementation, SuperShaper trains sub-networks that are competitive on GLUE tasks with BERT-base at the large end and with many compressed models (both hand-crafted and searched with NAS) at lower parameter counts.
In the size interval of 60-66M parameters, the model found with SuperShaper has a higher average accuracy on GLUE than larger models identified with many successful techniques such as LayerDrop, DistillBERT, Bert-PKD, miniLM, Tiny-BERT, BERT-of-Theseus, PD-BERT, ELM, and YOCO-Bert. 
Only NAS-BERT, with its much larger search space and knowledge distillation reports a higher accuracy by 1\%.

While SuperShaper finds competitive networks, we believe the key insight of SuperShaper is in the choice of the search space of shapes, the advantages it affords for deployability, and its implications for NAS in general.
The first advantage relates to generalized insights on {good shapes}.
With evolutionary algorithms (EAs) we search for sub-networks that are  accurate and meet given parameter constraints.
By analysing such networks, we identify heuristics of good shapes, which suggest a \textit{cigar-like} shape.
By applying these heuristics, we hand-craft sub-networks across a range of parameter counts.
These hand-crafted sub-networks match and often exceed the performance of networks searched with EAs.
Thus, Transformer shapes afford interpretable generalization of model compression across a range of parameter count constraints.
No existing NAS technique for super-training or more generally in NLP, demonstrate such generalization.
Second, we observe that the latency of networks as measured on  CPUs and GPUs is insensitive to the shape of the network: For a given parameter count that meets the latency requirement on a device, the shape of the network is a \textit{free variable} that can be optimized for accuracy. 
Thus, when deploying a model for a specific device, we need to compute the largest parameter count that is allowed by the latency constraint and then choose a network with a good shape for that parameter count.
We demonstrate that this procedure is effective across GPUs and CPUs.
In summary, SuperShaper is a simple yet effective technique to pretrain a range of task-agnostic language models to deploy appropriately sized NLP models on a range of devices.
More broadly, our findings suggest that Neural Architecture Search (NAS) can be performed with radically simpler design spaces and implementations, which generalize across tasks, parameter counts, and devices. 


\section{Related Work}
\label{sec:related}
\noindent 

Over the years, a number of solutions have been proposed for efficient deployment of language models. These can be broadly grouped into the following categories.

\subsection{Model Compression}
In the context of language models, model compression has been widely applied to reduce computational complexity. For instance, \cite{wang2019structured} and \cite{ma2019tensorized} propose low-rank approximations of weight matrices for faster matrix multiplications. A number of efforts prune either attention heads \cite{michel2019sixteen}, tokens \cite{goyal2020power,wang2021spatten,kim2021learned}, or even complete layers \cite{fan2019reducing,sajjad2020poor} to propose smaller and faster transformer models. Recent work on lottery-ticket hypothesis \cite{frankle2018lottery} applied to BERT models \cite{prasanna2020bert,chen2020lottery,chen2020earlybert,yu2019playing} has shown promise on finding good performing pruned models. Yet another successful approach for model compression is through quantization of weights and inputs to lower-precisions \cite{shen2020q,zafrir2019q8bert}. 

\subsection{Knowledge Distillation}
Knowledge distillation (KD) \cite{Hinton2015DistillingTK} aims to compress the knowledge from a large teacher model to a compact and fast student model. This can be further categorized into task-specific and task-agnostic knowledge distillation. Task specific distillation \cite{Tang2019DistillingTK, Turc2019WellReadSL, sun-etal-2019-patient, ijcai2020-0341} requires to first fine-tune the teacher model for each downstream task and then perform distillation,  whereas Task agnostic distillation \cite{Sanh2019DistilBERTAD, jiao2020tinybert, wang2020minilm, sun-etal-2020-mobilebert} aims to obtain a general student model for the language modeling task which can directly be finetuned for all downstream tasks. 

Generally, in KD the student models are trained with the soft targets obtained from the prediction layer of the teacher model with  KL divergence or Mean square error (MSE) used as loss functions. Recent works have also tried aligning the representations across the embedding layer outputs\cite{Sanh2019DistilBERTAD, jiao2020tinybert}, hidden states\cite{jiao2020tinybert, sun-etal-2020-mobilebert}, self attention distributions \cite{wang2020minilm, jiao2020tinybert}.


\subsection{Neural Architecture Search}
Neural Architecture Search \cite{zoph17iclr} aims to automate the design of neural networks by searching through a large space of network topologies and operators. The earlier efforts in computer vision \cite{zoph17iclr,liu2018darts} followed a costly bi-level optimization involving discrete outer loop search and continuous inner loop weight optimization. Soon, these were replaced with a more effective weight-sharing NAS approach \cite{cai2019once,yu2020bignas,wang2021attentivenas} that trains a large supernetwork followed by a careful search for optimal subnetworks. 

In NLP, \cite{wang2020hat} applied weight-sharing based NAS in the context of transformers to obtain efficient models for an end-to-end machine translation task. NAS is more challening for language modeling that involves task-agnostic pretraining and task-specific finetuning. Most NAS efforts in NLP focus on task-specific finetuning \cite{gao2021autobert,chen2adabert}. 
Recent contemporary efforts focus on the challenging search for task-agnostic models by building a super-network using design strategies such as block-wise search, progressive shrinking and stochastic gradient optimization \cite{xu2021bert,zhang2021you}.

In contrast, SuperShaper is a super-pretraining methodology to train a large number of task-agnostic and device-insensitive models in one-shot. The primary benefit of SuperShaper stems from its simple design-space. Many of the model-compression and knowledge-distillation efforts described here are complementary to SuperShaper and can be applied together to gain more benefits. The simplified design space also allows SuperShaper to effectively train a large number of models without requiring sophisticated training and search optimisations like other contemporary efforts \cite{xu2021bert,zhang2021you}. 

\section{SuperShaper: The methodology} 
\label{sec:supershaper}
\noindent 
In this section we describe the methodology used in SuperShaper focusing on the backbone network, methods for pretraining and fine-tuning, and for searching sub-networks.

\begin{figure}
    \centering
    \includegraphics[width=0.6\textwidth]{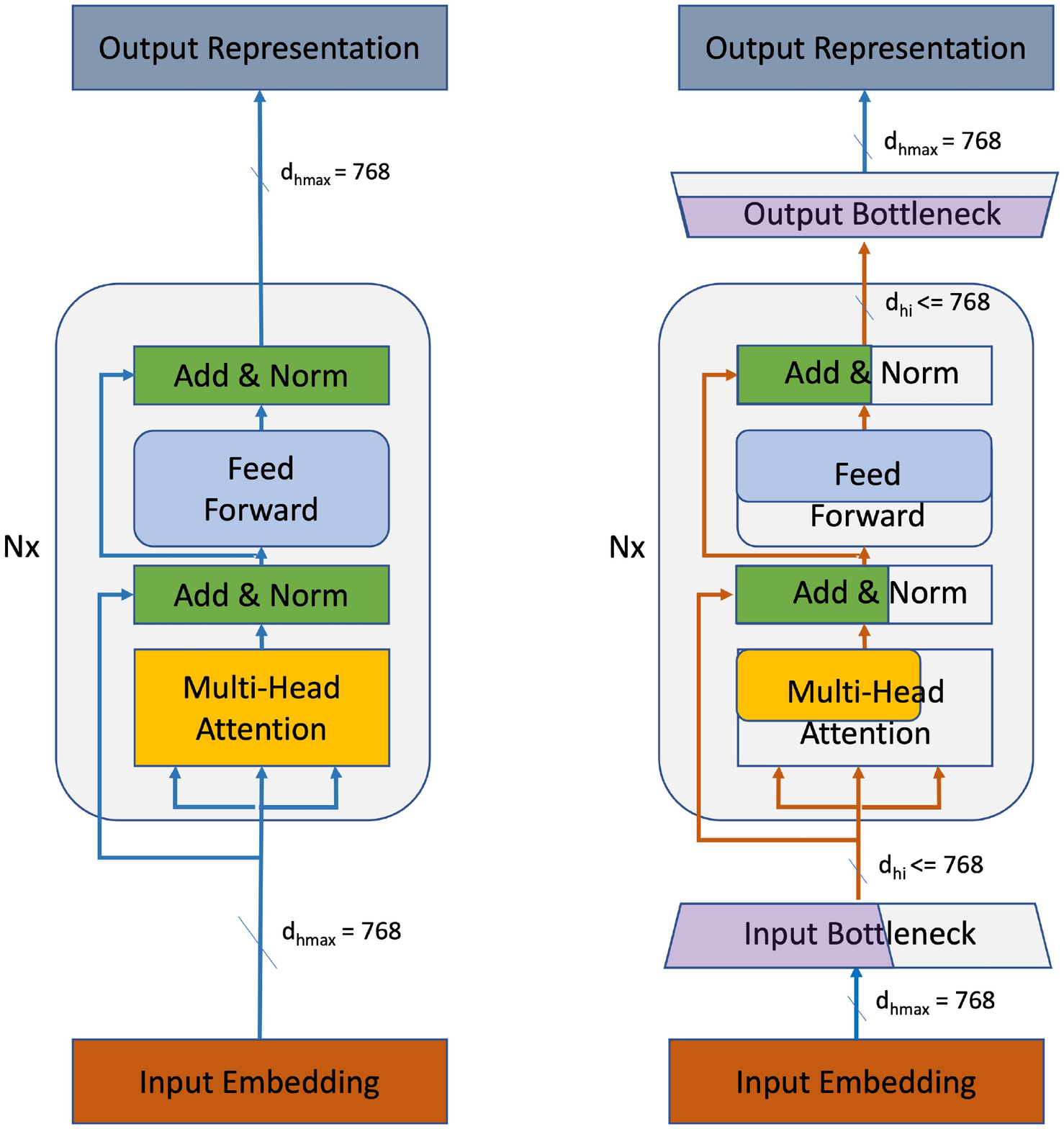}
    \caption{A Transformer layer in (a) BERT, and (b) Backbone used in SuperShaper with bottleneck matrices.}
    \label{fig:backbone}
\end{figure}


\subsection{SuperShaper Backbone}
\label{backbone}

A super pre-training procedure is characterized by a search space of networks.
For a Transformer layer in BERT-based networks, different search spaces have been studied, including the number of attention heads, the number of neurons in the FFNs, the number of encoder layers, the use of other operators like separable convolution, etc.
As discussed, SuperShaper searches over a single variable - the hidden dimensions for each layer. 
A standard Transformer layer \cite{vaswani2017attention} as visualized in \Cref{fig:backbone} (a) takes as input the representation per token in the sequence, passes it through a multi-headed self-attention module which is followed by an FFN module.
Residual connections and layer norms are present after either operation.
A characteristic of such a layer is the size of the representations of each input token, namely the hidden dimension denoted $d_h$.
The hidden dimension captures the amount of information coming into the layer but also characterizes the parameter and computational complexity of that layer. 
For a typical Transformer layer based on BERT-base with 12 attention heads (with $d_{attn}$ dimension), and FFNs (with $d_{ff}$ dimension), the number of parameters in a layer is given by $d_{h} \times (2 \times(2d_{attn} + d_{ff} + d_{h}))$ and the number of FLOPs is $d_h \times (4 \times(2d_{attn} + d_{ff} + d_{h})) + 2\times seq\_len \times d_{attn}$.

In a standard BERT-like encoder the hidden dimension $d_h$ of each layer is a constant, e.g., 768 for BERT-base.
But with SuperShaper, we would like to explore sub-networks where layers have different hidden dimensions.
The intuition behind this choice is that different layers may perform roles of varying importance. 
For instance, earlier layers manipulating the input embeddings and the final layers responsible for the output may require larger hidden dimensions.
To enable this layer-wise variability of hidden dimensions, we take inspiration from MobileBERT \cite{sun-etal-2020-mobilebert} which proposed a bottleneck layer to compress the parameter size of BERT.
Based on this, we modify the standard Transformer layer as shown in \Cref{fig:backbone} (b). 
The input and output of each layer are intermediated by multiplication with bottleneck matrices, which translate between the dimension of a token outside a layer (say 768) and the dimension of a token inside a layer (say 120).
With this change, each layer can have differently sized bottleneck matrices such that the hidden dimension varies across layers.

For pre-training, we want to create a family of networks with varying shapes.
To this end, we take inspiration from slicing matrices for elastic training \cite{cai2019once}.
Specifically, each bottleneck matrix is sized at the largest size of $768 \times 768$.
For sampling different sub-networks from this super-network, we \textit{slice} the bottleneck matrix. 
For instance, to reduce the dimension of a layer to 120, we slice the bottleneck matrix at the input from $768 \times 768$ to $768 \times 120$. 
In other words, we only retain the first 120 columns in the input bottleneck matrix, and equivalently only the first 120 rows in the bottleneck matrix at the output.
By so slicing the \textit{prefix} columns and rows of the matrix we can generate differently shaped sub-networks.
This slicing procedure is implemented with only about 20 lines of additional code in a PyTorch model as detailed in code snippet below. 
For a sliced matrix both forward and backward passes are similar to a standard neural network.

\begin{lstlisting}[
  language=python,
  title={PyTorch Code to Illustrate Slicing},
  captionpos=t,
  label={lst:pytorch}]
class CustomLinear(nn.Linear):
    def __init__(
        self, super_in_dim, super_out_dim, bias=True, uniform_=None, non_linear="linear"
    ):
    self.samples = {}
    ...
    def set_sample_config(self, sample_in_dim, sample_out_dim):
       sample_weight = weight[:, :sample_in_dim]
       sample_weight = sample_weight[:sample_out_dim, :]
       self.samples["weight"] = sample_weight  
       self.samples["bias"] = self.bias[..., : self.sample_out_dim]
        
    def forward(self, x):
       #override the Forward pass to use the sampled weights and bias
       return F.linear(x, self.samples["weight"], self.samples["bias"])

\end{lstlisting}

\subsection{Training with SuperShaper}

We formalize some notation of the networks.
We denote the SuperShaper backbone as $T$ and any sub-network sliced from the backbone as $T_S$ where $S$ is the \textit{shape} vector that represents the layer-wise hidden dimensions, $S_i$ for layer $i$.
The set of all possible values of $S$ denotes the design space $D$.
The smallest and largest sub-networks in $D$ are denoted as $T_{S^-}$ and $T_{S^+}$, respectively, while a random sub-network is denoted as $T_{S^r}$.
To evaluate how a sub-network $T_S$ has trained, we evaluate accuracy on Masked Language Modelling (MLM) task in terms of perplexity denoted $P(T_S)$.
Perplexity represents the uncertainty in predicting the representations of masked tokens in the MLM task. 








We now describe the training procedure for SuperShaper: From a given design space $D$, we \textit{sample} $n$ different shapes $S$ and obtain $T_S$ for each by the slicing technique described above.
This sampling can be performed in two ways: (a) uniform random sampling from $D$, and (b) random Sampling with \textit{sandwich rule} \cite{yu2019universally}, where in addition to (a) we also sample the largest and smallest sub-networks $T_{S^+}$ and $T_{S^-}$.
Sandwich rule has been shown to perform better for weight-sharing NAS in computer vision \cite{yu2019universally,yu2020bignas,wang2021attentivenas}. 
For language modelling, we study both sampling methods and report our findings in \Cref{sec:results}.
With the sampled sub-networks, gradient updates are computed and parameters are modified with a standard optimizer.
Note that the sub-networks share a large number of their parameters, in particular the earlier rows and columns of the bottleneck matrices.
Also matrices inside the layer (such as query, key, and value projection matrices) are shared. 
This parameter sharing is expected to enable generalization during training across the large space of sub-networks.
We evaluate and provide empirical evidence for such generalization in \Cref{sec:results}.

\subsection{Fine-tuning $T_S$ from SuperShaper}
The pre-training with SuperShaper is task-agnostic. 
To fine-tune a sampled sub-network $T_S$ for a specific task, several options exist.
First, we can sample $T_S$ and fine-tune it directly on the task.
Second, we can further pre-train $T_S$ individually and then fine-tune on the task.
Finally, we can randomly initialize the weights of $T_S$ and pre-train from scratch before fine-tuning. 
We compare these options by fine-tuning on NLU 8 tasks -- MNLI-m, QQP, QNLI, CoLA, SST-2, STS-B, RTE, MRPC -- from the GLUE benchmark \cite{wang2018glue}.
Across these tasks, we report the average performance for the three options as \gdirect, \gpartial, and \gscratch, respectively.

%

\subsection{Searching for optimal shapes}
Once we have super pre-trained with a design space $D$, we can sample and deploy $T_S$ for any $S$, which can then be fine-tuned by methods described in the previous subsection. 
The set of sub-networks can be large: For instance, for a design space with 7 options for hidden dimensions at 12 layers, we have almost 14 billion sub-networks.
The search question is to find an optimal shape from amongst this large set which meets specific constraints on accuracy, parameter count, or latency on devices.
We adopt Evolutionary Algorithm (EA) from \cite{real2017large} as a generic optimization technique, which starts with a population of solutions and over generations performs genetic operations such as mutation and crossover to create new solutions.
At each generation the fittest solutions are retained based on defined metrics of fitness.
For SuperShaper, the representation of sub-networks and genetic operations are natural since a sub-network is specified by a shape vector, equivalent to a genetic representation of an solution.
For the fitness metrics, we use perplexity on language modelling and latency on a device.
However, given that computing these metrics for thousands of candidate solutions would be expensive, we use trained predictors to approximate perplexity and latency as been studied elsewhere in NAS \cite{cai2019once,ganesan2020case}. 

While EA with fitness predictors can search for sub-networks, the most desirable setting is to find sub-networks by applying a set of heuristics to decide the shape of each layer.
We propose a technique to discover such heuristics and then use is to identify sub-networks for varying parameter count constraints.
We report results on how these compare against EAs in \Cref{sec:results}.

\section{Experimental Setup and Results}
\label{sec:results}
\noindent 
In this section, we detail the experimental setup and report a range of experimental findings to evaluate SuperShaper. 

\subsection{Experimental Setup}
We first describe the pre-training, fine-tuning, and experimental setup used in all our experiments. 

\noindent\textbf{Design space} We slice the bottleneck matrix to produce Transformer layers of varying hidden dimensions.
Our design space for hidden dimensions is \{120, 240, 360, 480, 540, 600, 768\}, which creates a design space $D$ of $7^{12}$ or about 14 billion sub-networks.

\noindent\textbf{Super-pretraining}
We initialize our backbone with BERT-base-cased model trained on Wikipedia and BookCorpus. 
We then super pre-train the backbone using Masked language modeling over the C4 RealNews dataset \cite{raffel2019exploring} with batch size 2048, max sequence length 128, for 175K steps (or 26 epochs) on 8 A100 GPUs. The complete set of hyperparameters used is shown in \Cref{tab:mlmhyperparams}

\begin{table}[th!]
\centering
\begin{tabular}{@{}l|c@{}}
\toprule
Data & C4/RealNews \\
Max sequence length & 128 \\
Batch size & 2048 \\
Peak learning rate & 2e-5\\
Number of steps & 175K \\
Warmup steps & 10K \\
Hidden dropout & 0 \\
GeLU dropout & 0 \\
Attention dropout & 0 \\
Learning rate decay & Linear \\
Optimizer & AdamW \\
Adam $\epsilon$ & 1e-6 \\
Adam $(\beta_1, \beta_2)$ & (0.9, 0.999) \\
Weight decay & 0.01 \\ 
Gradient clipping & 0 \\
\bottomrule
\end{tabular}
\caption{Hyperparameters for MLM super pre-training on C4 RealNews. Super pre-training was done on 8 A100 GPUs}
\label{tab:mlmhyperparams}
\end{table}

\noindent\textbf{Fine tuning.}
Similar to \cite{xu2021bert}, we evaluate the effectiveness of SuperShaper by pre-training all our compressed models from scratch and later fine-tuning them on 8 GLUE tasks. 
We report performance metrics on the \texttt{dev} version of the benchmark. For RTE, MRPC and STS-B, we start with a model fine-tuned on MNLI similar to \cite{liu2019roberta, xu2021bert}.
For metrics, we report Matthews correlation for CoLA \cite{wang2018glue}, Spearman correlation for STS-B \cite{wang2018glue} and accuracy for all other tasks. 
For MNLI-m \cite{wang2018glue}, we report accuracy on the matched set. 
Following \cite{devlin2019bert, xu2021bert, zhang2021you, ludyna2020}, we also exclude the problematic WNLI dataset. 
For all the datasets in GLUE, we use the official train and dev splits and download the datasets from HuggingFace datasets\footnote{https://huggingface.co/datasets/glue}. The complete set of hyperparameters for fine-tuning can be found in \Cref{tab:finetuningparams}.

\begin{table}[th!]
\centering
\begin{tabular}{@{}l|c|c@{}}
\toprule
 & CoLA & Other GLUE tasks\\ \midrule
Batch size & {\{16, 32\}} & 32 \\
Weight decay & {\{0, 0.1\}} & 0 \\
Warmup steps & {\{0, 400\}} & 0 \\
\midrule
Max sequence length & \multicolumn{2}{c}{128} \\
Peak learning rate & \multicolumn{2}{c}{5e-5}\\
Number of epochs & \multicolumn{2}{c}{10} \\
Hidden dropout & \multicolumn{2}{c}{0} \\
GeLU dropout & \multicolumn{2}{c}{0} \\
Attention dropout & \multicolumn{2}{c}{0} \\
Learning rate decay & \multicolumn{2}{c}{Linear} \\
Optimizer & \multicolumn{2}{c}{AdamW} \\
Adam $\epsilon$ & \multicolumn{2}{c}{1e-6} \\
Adam $(\beta_1, \beta_2)$ & \multicolumn{2}{c}{(0.9, 0.999)} \\
Gradient clipping & \multicolumn{2}{c}{0} \\
\bottomrule
\end{tabular}
\caption{Hyperparameters for fine-tuning on GLUE}
\label{tab:finetuningparams}
\end{table}

\noindent\textbf{Evolutionary Algorithm (EA).} 
For EA, we adapt the algorithm presented in \cite{real2017large}.
We choose a population size of $100$, mutation probability of $0.4$, and the ratio of parent size to mutation or crossover size as $1$. 
We bound the search algorithm to $300$ iterations. 

\noindent\textbf{Fitness Predictors.}
For perplexity predictor, we randomly sample $10,000$ sub-networks and evaluate their perplexity as measured on C4-RealNews dataset. 
We use this dataset to build the predictor based on XGBoost model \cite{chen2015xgboost}.
For latency predictor, we sample $1,000 - 4,000$ sub-networks and evaluate their latency on the chosen device. 
We again train a XGBoost model to predict latency from this dataset.
We consider 3 devices - 2 GPUs 1080Ti and K80, and a server class single-core Xeon CPU. 
The GPUs were evaluated with a batch size of 128 while the CPU was evaluated with batch size 1. 


\subsection{Pre-training with SuperShaper}
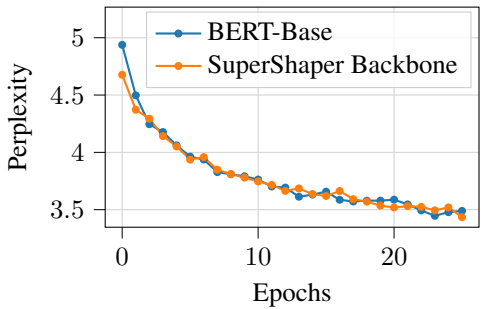
\begin{figure}[!h]
    \centering
\begin{tikzpicture}

\definecolor{color0}{rgb}{0.12156862745098,0.466666666666667,0.705882352941177}
\definecolor{color1}{rgb}{1,0.498039215686275,0.0549019607843137}
\definecolor{color2}{rgb}{0.172549019607843,0.627450980392157,0.172549019607843}

\begin{axis}[
legend cell align={left},
legend style={fill opacity=0.8, draw opacity=1, text opacity=1, draw=white!80!black},
tick align=outside,
tick pos=left,
x grid style={white!82.7450980392157!black},
xlabel={Epochs},
xmajorgrids,
xmin=-1.25, xmax=26.25,
xtick style={color=black},
y grid style={white!82.7450980392157!black},
ylabel={Perplexity},
ymajorgrids,
ymin=3.34691982526356, ymax=5.26705351023564,
ytick style={color=black},
height=4.5cm, width=6.5cm
]
\addplot [thick, color0, mark=*, mark size=1.1, mark options={solid}]
table {%
0 4.93800020217896
1 4.49700021743774
2 4.24528074264526
3 4.17742109298706
4 4.06135749816895
5 3.96226072311401
6 3.93767786026001
7 3.8277633190155
8 3.80919480323792
9 3.79116535186768
10 3.76299524307251
11 3.70206522941589
12 3.69290709495544
13 3.6130952835083
14 3.63144588470459
15 3.65731954574585
16 3.58637976646423
17 3.57128119468689
18 3.57842612266541
19 3.57903027534485
20 3.58745074272156
21 3.54599595069885
22 3.49378395080566
23 3.44647312164307
24 3.47842383384705
25 3.48928260803223
};
\addlegendentry{BERT-Base}
\addplot [thick, color1, mark=*, mark size=1.1, mark options={solid}]
table {%
0 4.67690944671631
1 4.37168550491333
2 4.29375314712524
3 4.14207220077515
4 4.05190134048462
5 3.936847448349
6 3.95853805541992
7 3.84874677658081
8 3.81068134307861
9 3.77950739860535
10 3.74651575088501
11 3.71655106544495
12 3.66392207145691
13 3.68517088890076
14 3.63624572753906
15 3.61902379989624
16 3.66286110877991
17 3.59161591529846
18 3.56948637962341
19 3.53516697883606
20 3.51926875114441
21 3.5311815738678
22 3.52472019195557
23 3.49411749839783
24 3.52000975608826
25 3.43419861793518
};
\addlegendentry{SuperShaper Backbone}
\end{axis}
\end{tikzpicture}
\label{fig:perpx_bottleneck}
\caption{The perplexity trajectory of BERT-base and SuperShaper backbone on MLM task for C4 Real News}
\end{figure}
\input{Images/loss_perpx_dup}
\subsubsection{Training the backbone.}
We first pre-train our proposed backbone network that is based on BERT-base but with additional bottlenneck matrices on the C4-RealNews dataset.
We initialize the bottleneck structures with identity weights and zero-bias. 
The corresponding plot of perplexity on language modelling is for BERT-base and our backbone are shown in \Cref{fig:perpx_bottleneck}. 
The SuperShaper backbone mimics the performance of BERT-base thereby providing a good starting point for pre-training of multiple sub-networks with SuperShaper.


\subsubsection{Effect of sub-network sampling rule.}

\begin{figure}[!ht]
    \centering
    \includegraphics[width=0.45\textwidth]{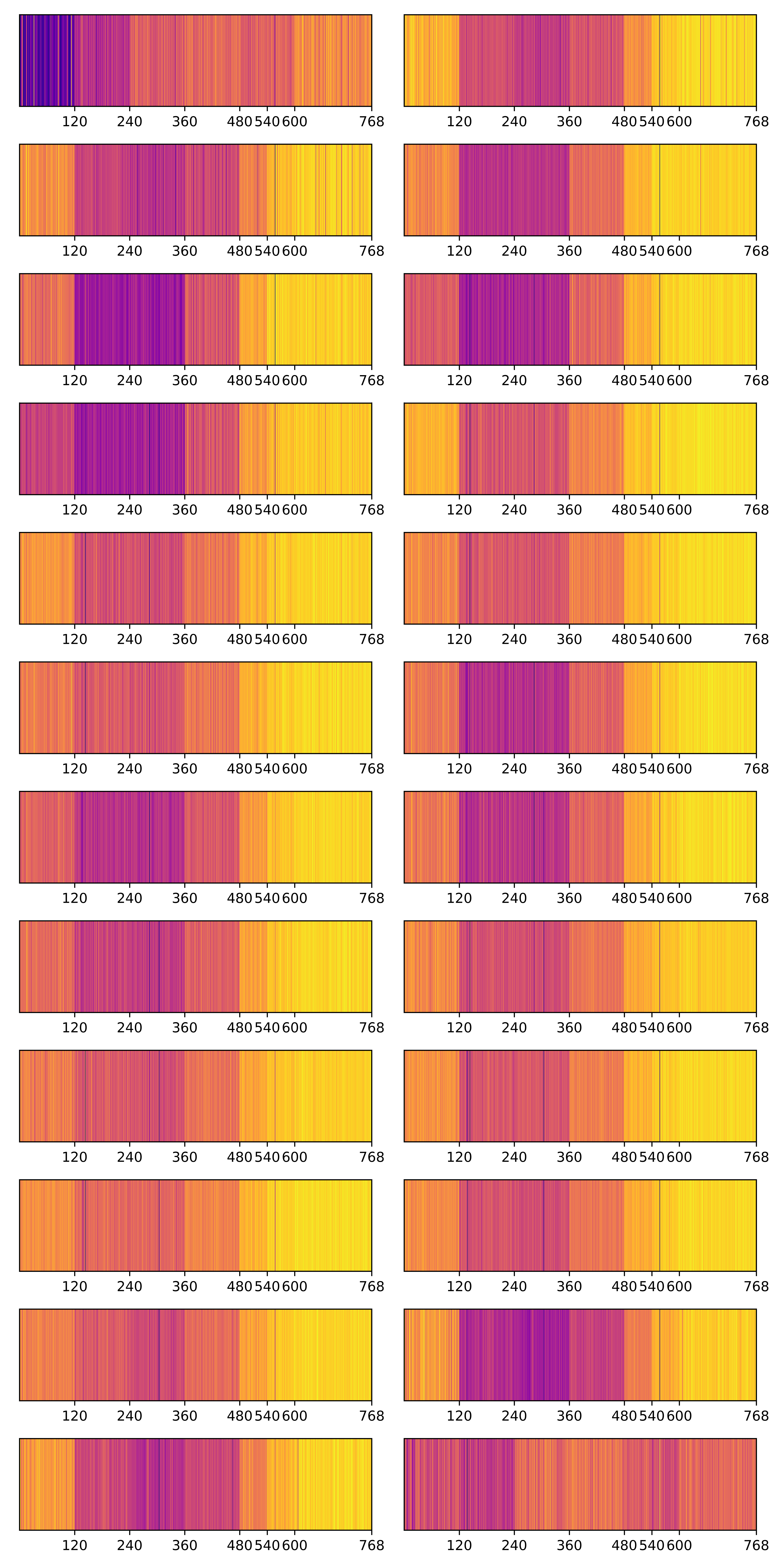}
    \caption{Visualization of input and output Bottleneck matrix diagonals for all the 12 layers.}
    \label{fig:bottleneck_all_viz}
\end{figure}

In computer vision, sandwich rule is widely used in the context of weight-sharing NAS \cite{yu2019universally}.
We super pre-train the trained backbone network with the sandwich rule. 
The corresponding loss trajectory for largest, smallest, and randomly sampled sub-networks are shown in \Cref{fig:loss_perpx}(a).
Clearly, the larger network has a lower perplexity, but the super pre-training ensures that a range of networks are simultaneously trained on the MLM task.
Specifically, randomly sampled subnetworks shown as $T_{S^r}$ even though not sampled as frequently as the smallest subnetwork, have a lower perplexity.
This provides evidence of generalization during super pre-training.

We now compare the sandwich sampling rule with fully randomised sampling. 
We plot the perplexity of 4 networks: the largest, smallest, and two other intermediate networks in \Cref{fig:loss_perpx}(b)-(e).
Sandwich sampling always samples the largest and smallest and thus the perplexity on these networks is significantly lower with sandwich sampling than random sampling.
On the other two networks, sandwich sampling is able to tightly match the perplexity obtained with random sampling, even though stochastically sandwich rule has fewer gradient updates to these two networks than random sampling.
This suggests that sandwhich sampling effectively combines good extremum sub-networks with reasonably good intermediate sub-networks.
In all subsequent experiments, we use sandwich sampling.

\subsubsection{Visualizing bottleneck matrices.}
We initialize the bottleneck matrices to identity.
After super pre-training, we visualize these matrices to understand the role of sliced training of sub-networks.
We take the softmax of the principal diagonal of the two bottleneck matrices, and plot them for all layers in \Cref{fig:bottleneck_all_viz}.
We clearly observe that the entries show a banded pattern with boundaries at the shapes in our design space: 120, 240, 360, 480, 540, 600, and 768.
This implies that super pre-training learns different linear projections of 768 dimensional input representation to the chosen hidden dimensions.
 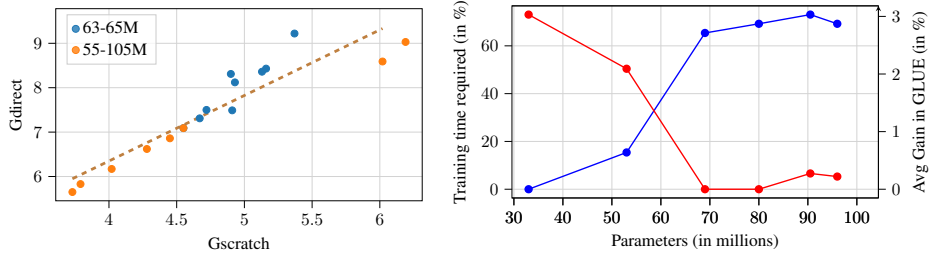
\begin{figure*}[!ht]
     \centering
     \scalebox{0.65}{
     \begin{tabular}{cc}
      \begin{tikzpicture}

\definecolor{color0}{rgb}{0.12156862745098,0.466666666666667,0.705882352941177}
\definecolor{color1}{rgb}{1,0.498039215686275,0.0549019607843137}

\begin{axis}[
legend cell align={left},
legend style={
  fill opacity=0.8,
  draw opacity=1,
  text opacity=1,
  at={(0.03,0.97)},
  anchor=north west,
  draw=white!80!black
},
tick align=outside,
tick pos=left,
x grid style={white!82.7450980392157!black},
xlabel={Gscratch},
xmajorgrids,
xmin=3.607, xmax=6.313,
xtick style={color=black},
y grid style={white!82.7450980392157!black},
ylabel={Gdirect},
ymajorgrids,
ymin=5.45334640043758, ymax=9.77972559081074,
ytick style={color=black},
height=5.5cm, width=9cm
]
\addplot [draw=color0, fill=color0, mark=*, only marks]
table{%
x  y
4.55 7.09
4.67 7.31
4.91 7.49
4.72 7.5
4.93 8.12
4.9 8.31
5.13 8.36
5.16 8.43
5.37 9.22
};
\addlegendentry{63-65M}
\addplot [draw=color1, fill=color1, mark=*, only marks]
table{%
x  y
6.19 9.03
6.02 8.59
4.55 7.09
4.45 6.86
4.28 6.62
4.02 6.17
3.79 5.83
3.73 5.65
};
\addlegendentry{55-105M}
\addplot [ultra thick, brown, dashed, forget plot]
table {%
3.73000001907349 5.94954872131348
4.55000019073486 7.1607232093811
6.01999998092651 9.33197498321533
};

\end{axis}
\end{tikzpicture}  & \begin{tikzpicture}

\begin{axis}[
tick align=outside,
tick pos=left,
x grid style={white!82.7450980392157!black},
xlabel={Parameters (in millions)},
xmajorgrids,
xmin=29.6, xmax=104.4,
xtick style={color=black},
y grid style={white!82.7450980392157!black},
ylabel={Training time required (in \%)},
ymajorgrids,
ymin=-3.653846154, ymax=76.730769234,
ytick style={color=black},
height=5.5cm, width=9cm
]
\addplot [thick, blue, mark=*, mark size=2, mark options={solid}]
table {%
33 0
53 15.384614944458
69 65.3846130371094
80 69.2307662963867
90.5 73.0769195556641
96 69.2307662963867
};
\end{axis}

\begin{axis}[
axis y line=right,
tick align=outside,
x grid style={white!69.0196078431373!black},
xmin=29.6, xmax=104.4,
xtick pos=left,
xtick style={color=black},
y grid style={white!69.0196078431373!black},
ylabel={Avg Gain in GLUE (in \%)},
ymin=-0.1515625, ymax=3.1828125,
ytick pos=right,
ytick style={color=black},
yticklabel style={anchor=west},
height=5.5cm, width=9cm
]
\addplot [thick, red, mark=*, mark size=2, mark options={solid}]
table {%
33 3.03125
53 2.09124994277954
69 0
80 0
90.5 0.273750066757202
96 0.219874978065491
};
\end{axis}
\end{tikzpicture}
     \end{tabular}
     }
     \caption{
     (a) SuperShaper provides a fast and accurate proxy for sub-network perplexity, and (b) \gpartial~inherited sub-networks only require fraction of pre-training cost (in blue) i.e. 1.3-6.6x reduction to reach optimum. This comes at a higher average gain in GLUE score (in red).}
     \label{fig:supernet-characteristics}
 \end{figure*}




\subsubsection{Effectiveness of super pre-training.}
We ask two questions towards evaluating the effectiveness of super pre-training: 
(a) Is the relative performance of sampled sub-networks on the MLM perplexity (\gdirect) correlated with performance of the same sub-networks when pre-trained individually from scratch (\gscratch)?, and 
(b) Does the super pre-training afford sub-networks an advantage when being fine-tuned for tasks?
For the first question, we sample a set of sub-networks $T_S$ of both varying (33-96M) and similar (63-65M) parameter counts, and plot \gdirect~and \gscratch~in \Cref{fig:supernet-characteristics} (a). 
We notice that \gdirect~and \gscratch~are highly correlated with a Spearman correlation coefficient of 0.954. 
This implies that the sub-network's measured MLM perplexity after super pre-training is a good proxy for final performance. 
We also observe that networks sampled at the similar parameter count (63-65M) have varying performance suggesting the sensitive role of shape in accuracy. 

For studying the second question, we pre-train and then fine-tune the varying parameter count sub-networks (33-96M) in two ways (a) by retaining the weights learnt during super pre-training (\gpartial), and (b) starting with random initialization \gscratch. 
We plot two quantities in \Cref{fig:supernet-characteristics} (b): the amount of pre-training time saved with (a) and the additional GLUE score obtained with (a). 
We observe that models with fewer parameters (30-50M) show significant savings in the pre-training time (up to $6.6\times$) and simultaneously benefit from improved GLUE accuracy (up to $3\%$).
The gains on both axes for larger models are smaller.
This suggests that smaller models whose parameters receive more weight updates due to sharing of the earlier rows and columns across sub-networks benefit more from super pre-training.
This is encouraging because most effort in deployability is concerned with models of smaller size.



\begin{table*}[!ht]
\centering
\scalebox{1.0}{
\scriptsize
\begin{tabular}{lrrrrrrrrrrr}\toprule
\textbf{Model} &\textbf{Params} &\textbf{MNLI-m} &\textbf{QQP} &\textbf{QNLI} &\textbf{CoLA} &\textbf{SST-2} &\textbf{STS-B} &\textbf{RTE} &\textbf{MRPC} &\textbf{Average GLUE} \\\cmidrule{1-11}
\textbf{BERT-Base (from NAS-BERT)} & 110M & 85.2 & 91 & 91.3 & 61 & 92.9 & 90.3 & 76 & 87.7 & 84.4 \\\cmidrule{1-11}
\textbf{BERT-Base (from DistilBERT)}  & 110M & 86.7 & 89.6 & 91.8 & 56.3 & 92.7 & 89 & 69.3 & 88.6 & 83 \\\cmidrule{1-11}
\textbf{SuperShaper-Base} & 96M & 83.9 & 90.86 & 90.92 & 56.58 & 92.89 & 88.3 & 77.98 & 88.48 & 83.7 \\\midrule
\bottomrule
\end{tabular}
}
\caption{Comparing SuperShaper with BERT-Base models.}
\label{tab:super-base}
\end{table*}

\begin{table*}[!ht]\centering
\scalebox{0.8}{
\begin{tabular}{lrrrrrrrrrrr}
\toprule
\textbf{Model} &\textbf{Params} &\textbf{MNLI-m} &\textbf{QQP} &\textbf{QNLI} &\textbf{CoLA} &\textbf{SST-2} &\textbf{STS-B} &\textbf{RTE} &\textbf{MRPC} &\textbf{Average GLUE} \\\midrule
LayerDrop \cite{fan2019reducing}  &66M &80.7 &88.3 &88.4 &45.4 &90.7 &- &65.2 &85.9 & 77.8 \\\cmidrule{1-11}
DistilBERT \cite{sanh2019distilbert} &66M &82.2 &88.5 &89.2 &51.3 &91.3 &86.9 &59.9 &87.5 &79.6 \\\cmidrule{1-11}
Bert-PKD \cite{sun2019patient} &66M &81.5 &70.7 &89.0 &- &92.0 &- &65.5 &85.0 & 80.6 \\\cmidrule{1-11}
MiniLM \cite{wang2020minilm} &66M &84.0 &91.0 &91.0 &49.2 &92.0 &- &71.5 &88.4 &81.0 \\\cmidrule{1-11}
Ta-TinyBert \cite{jiao2020tinybert} &67M &83.5 &90.6 &90.5 &42.8 &91.6 &86.5 &72.2 &88.4 &80.8 \\\cmidrule{1-11}
Tiny-BERT \cite{jiao2020tinybert} &66M &84.6 &89.1 &90.4 &51.1 &93.1 &83.7 &70.0 &82.6 &80.6 \\\cmidrule{1-11}
BERT-of-Theseus \cite{xu2020bert} &66M &82.3 &89.6 &89.5 &51.1 &91.5 &88.7 &68.2 &- &80.1 \\\cmidrule{1-11}
PD-BERT \cite{turc2019well} &66M &82.5 &90.7 &89.4 & - & 91.1 & - &66.7 &84.9 &84.2 \\\cmidrule{1-11}
ELM \cite{jiao2021improving} &60M &84.2 &91.1 &90.8 &54.2 &92.7 &88.9 &72.2 &89.0 &82.9 \\
\toprule
NAS-BERT$\ast$ \cite{xu2021bert} &60M &83.3 &90.9 &91.3 &55.6 &92.0 &88.6 &78.5 &87.5 &83.5 \\\cmidrule{1-11}
DynaBERT$\dagger$ \cite{ludyna2020} &60M &84.2 &91.2 &91.5 &56.8 &92.7 &89.2 &72.2 &84.1 &82.8 \\\cmidrule{1-11}
YOCO-bert \cite{zhang2021you} & ~59-67M &82.6 &90.5 &87.2 &59.8 &92.8 & - &72.9 &90.3 & 82.3 \\\cmidrule{1-11}
SuperShaper evo-search (ours) & 63M & 82.2 &90.2 &88.1 &53.0 &91.9 &87.6 &79.1 &89.5 & 82.7 \\\cmidrule{1-11}
SuperShaper heuristic-shaped (ours) & 61M & 82 &90.3 &88.4 &52.6 &91.6 &87.8 &77.6 &86.5 & 82.1 \\
\bottomrule
\end{tabular}
}
\caption{Comparison of SuperShaper with 60-67M parameter constraint models on development set of GLUE. $\dagger$ indicates models trained with data augmentation, $\ast$ indicates model trained without knowledge distallation in the fine-tuning stage }\label{tab:supershaper-best}
\end{table*}

\begin{table*}[!ht]\centering
\scriptsize
\begin{tabular}{lrrrrrrrrrrrr}\toprule
\textbf{Params (M)} &\textbf{\gpartial} &\textbf{\gscratch} &\textbf{MNLI-m} &\textbf{QQP} &\textbf{QNLI} &\textbf{CoLA} &\textbf{SST-2} &\textbf{STS-B} &\textbf{RTE} &\textbf{MRPC} &\begin{tabular}{@{}c@{}}\textbf{Average} \\ \textbf{GLUE} \end{tabular} \\\cmidrule{1-12}
33 &10.82 &12.44 &73.45 &84.71 &80.52 &10.27 &85.32 &82.65 &65.70 &82.11 &70.59 \\\cmidrule{1-12}
53 &8.59 &6.02 &79.40 &89.51 &86.38 &33.85 &89.11 &86.66 &68.23 &84.56 &77.21 \\\cmidrule{1-12}
63 &7.09 &4.55 &82.23 &90.18 &88.05 &53.00 &91.86 &87.63 &79.06 &89.46 &82.68 \\\cmidrule{1-12}
69 &6.62 &4.28 &82.74 &90.45 &89.54 &54.98 &91.28 &88.42 &77.98 &87.75 &82.89 \\\cmidrule{1-12}
80 &6.17 &4.02 &83.05 &90.56 &89.22 &54.87 &93.10 &88.46 &80.14 &87.75 &83.39 \\\cmidrule{1-12}
90.5 &5.83 &3.79 &83.06 &90.51 &88.72 &58.87 &91.51 &88.47 &77.26 &88.97 &83.42 \\\cmidrule{1-12}
96 &5.65 &3.73 &83.90 &90.86 &90.92 &56.58 &92.89 &88.30 &77.98 &88.48 &83.74 \\\midrule
\bottomrule
\end{tabular}
\caption{Performance of best models from parameter-constrained evolutionary search}\label{tab:pareto_performance}
\end{table*}
\begin{table*}[!ht]\centering
\scriptsize
\begin{tabular}{lrrrrrrrrrrrrrrrr}\toprule
Shapes &Params (M) &G\_direct &G\_scratch &L1 &L2 &L3 &L4 &L5 &L6 &L7 &L8 &L9 &L10 &L11 &L12 \\\cmidrule{1-16}
EvoSearch 1 &65 &6.86 &4.45 &480 &360 &360 &240 &240 &360 &480 &480 &360 &480 &540 &540 \\\cmidrule{1-16}
Evo Search 2 &63 &7.09 &4.55 &480 &240 &360 &240 &540 &480 &360 &360 &360 &360 &540 &480 \\\cmidrule{1-16}
Lower Triangle &64 &7.31 &4.67 &120 &120 &240 &240 &360 &360 &360 &480 &540 &540 &600 &768 \\\cmidrule{1-16}
Random &64 &7.49 &4.91 &480 &360 &360 &540 &480 &540 &360 &480 &540 &120 &360 &540 \\\cmidrule{1-16}
Rectangle &58 &7.5 &4.72 &360 &360 &360 &360 &360 &360 &360 &360 &360 &360 &360 &360 \\\cmidrule{1-16}
Inverted Diamond &65 &8.12 &4.93 &768 &600 &360 &240 &240 &120 &120 &240 &240 &360 &600 &768 \\\cmidrule{1-16}
Bottle &64 &8.31 &4.9 &120 &120 &120 &120 &120 &120 &600 &600 &600 &600 &600 &768 \\\cmidrule{1-16}
Diamond &64 &8.36 &5.13 &120 &240 &360 &480 &480 &540 &768 &540 &480 &360 &240 &120 \\\cmidrule{1-16}
Upper Triangle &64 &8.43 &5.16 &768 &600 &540 &540 &480 &360 &360 &360 &240 &240 &120 &120 \\\cmidrule{1-16}
Inverted Bottle &64 &9.22 &5.37 &768 &600 &600 &600 &600 &600 &120 &120 &120 &120 &120 &120 \\\midrule
\bottomrule
\end{tabular}
\caption{Hidden dimensions of templatized shapes and their corresponding perplexities for \gscratch~and \gdirect.}
\label{tab:templatized}
\end{table*}
 \begin{table*}[!ht]\centering
 \scriptsize
 \begin{tabular}{lrrrrrr}\toprule
 \textbf{Evo-search parameters} &\textbf{\gdirect} &\textbf{Parameter range} &\textbf{Number of networks} &\textbf{Spearman Correlation} &\textbf{Pearson Correlation} \\\cmidrule{1-6}
 53   & 8.59 & 52-54 & 54   & 71.03 & 67.95 \\\cmidrule{1-6}
 63   & 7.09 & 62-64 & 704  & 80.34 & 82.52 \\\cmidrule{1-6}
 65   & 6.86 & 63-65 & 862  & 69.47 & 71.34 \\\cmidrule{1-6}
 69   & 6.62 & 68-70 & 1065 & 47.22 & 52.06 \\\cmidrule{1-6}
 80   & 6.17 & 79-81 & 486  & 72.32 & 67.34 \\\cmidrule{1-6}
 90.5 & 5.83 & 89-91 & 47   & 56.8  & 54.67 \\\cmidrule{1-6}
 96   & 5.65 & 95-97 & 6    & 65.71 & 69.65 \\\midrule
 \bottomrule
 \end{tabular}
 \caption{Shape difference positively correlates with \gdirect difference across a wide parameter range}
 \label{tab:l2norm}
 \end{table*}

\subsection{Comparing sub-networks with other methods}

\subsubsection{Comparing with BERT-base.}
As a first baseline, we search for a SuperShaper-Base model with EA with a constraint of 100M parameters and obtain a model with 96M parameters. 
This model is comparable against uncompressed BERT-base model which has 110M parameters.
We compare the GLUE performance \gscratch~of SuperShaper-Base with two of the top reported results on BERT-Base \cite{xu2021bert, sanh2019distilbert}.
From \Cref{tab:super-base}, we find that the average GLUE score across the two reported BERT-base baselines (84.4\% and 83\%) is similar to SuperShaper-Base (83.7\%).
Thus, SuperShaper-Base performs competitively with the uncompressed BERT-base with fewer parameters (96M vs 110M).

\subsubsection{Comparing with compressed models}
We now compare against state-of-the-art compressed models either hand-crafted or found by NAS algorithms (see \Cref{tab:supershaper-best}).
Since several of these models are in the range of 60-67M, we search for a sub-network from SuperShaper with a parameter constraint of 66M.
The task-wise performance of the obtained sub-network is reported in \Cref{tab:supershaper-best}.
SuperShaper outperforms many prominent hand-crafted or compressed networks proposed over the last two years by a significant margin. 
Across NAS-based methods, SuperShaper performs competitively despite a much simpler design space.
Only NAS-BERT reports a higher accuracy, which may be attributed to the use of novel operators such as separable convolution in the design space.
Also, NAS-BERT and DynaBERT use knowledge distillation and data augmentation.
These methods are orthogonal to shaping and can be combined with our approach.
In summary, we establish that SuperShaper with a simple design space and efficient super pre-training implementation performs competitively in compressing models to a given parameter count.


\begin{figure}[!ht]
    \centering
    \includegraphics[height=6cm,bb=0 0 800 600]{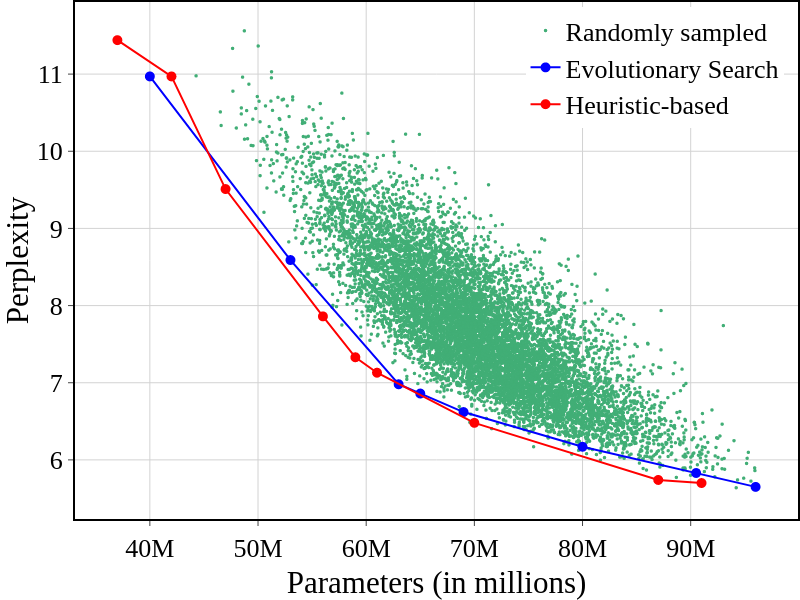}
    \caption{Evolutionary search finds accurate models, while models based on simple heuristics perform competitively.}
    \label{fig:pareto_heuristic}
\end{figure}

\subsubsection{Models for various parameter count constraints.}
We now apply EA to search for sub-networks at varying parameter count, ranging from 40 to 110M.
To understand the effectiveness of EA search, we sample 10,000 random sub-networks and compute their perplexity.
We then plot these points along with the networks searched by EA in \Cref{fig:pareto_heuristic}.
First, we observe that sub-network's shape critically affects accuracy.
Second, EA effectively searches for accurate networks across the parameter range.
We report GLUE scores for these networks in \Cref{tab:pareto_performance}. 



\subsection{Shape analysis of Super-Networks.}
In contrast to other NAS techniques, the design space of SuperShaper is interpretable - the network shape.
We can thus ask the question: Are there \textit{good shapes} across sizes?

\subsubsection{Models with templated shapes.}
We evaluate the performance of the following templated shapes in the 63-65M parameter range: lower triangle, upper triangle, rectangle, diamond, inverted diamond, bottle, and inverted bottle.
Details of the hidden dimensions and sub-network perplexity for each network are reported in \Cref{tab:templatized} 
We observe that lower triangle has the lowest perplexity (7.31) while inverted bottle (9.22) has the highest.
This wide range reiterates that shape sensitively affects performance.
Further, we observe that larger number of parameters towards the later layers benefits model performance.



\begin{figure*}[!ht]
\centering
\begin{tikzpicture}

\definecolor{color0}{rgb}{0.933333333333333,0.509803921568627,0.933333333333333}

\begin{groupplot}[group style={group size=4 by 1}]
\nextgroupplot[
tick align=outside,
tick pos=left,
x grid style={white!82.7450980392157!black},
xlabel={Features},
xmajorgrids,
xmin=0.23, xmax=12.77,
xtick style={color=black},
y grid style={white!82.7450980392157!black},
ylabel={Importance Scores (in \%)},
ymajorgrids,
ymin=0, ymax=31.5315,
ytick style={color=black},
height=3cm, width=4cm
]
\draw[draw=none,fill=green!50.1960784313725!black] (axis cs:0.8,0) rectangle (axis cs:1.2,5.95);
\draw[draw=none,fill=green!50.1960784313725!black] (axis cs:1.8,0) rectangle (axis cs:2.2,4.152);
\draw[draw=none,fill=green!50.1960784313725!black] (axis cs:2.8,0) rectangle (axis cs:3.2,4.196);
\draw[draw=none,fill=green!50.1960784313725!black] (axis cs:3.8,0) rectangle (axis cs:4.2,3.951);
\draw[draw=none,fill=green!50.1960784313725!black] (axis cs:4.8,0) rectangle (axis cs:5.2,6.3421);
\draw[draw=none,fill=green!50.1960784313725!black] (axis cs:5.8,0) rectangle (axis cs:6.2,6.756);
\draw[draw=none,fill=green!50.1960784313725!black] (axis cs:6.8,0) rectangle (axis cs:7.2,6.63);
\draw[draw=none,fill=green!50.1960784313725!black] (axis cs:7.8,0) rectangle (axis cs:8.2,6.81);
\draw[draw=none,fill=green!50.1960784313725!black] (axis cs:8.8,0) rectangle (axis cs:9.2,7.1);
\draw[draw=none,fill=green!50.1960784313725!black] (axis cs:9.8,0) rectangle (axis cs:10.2,7.74);
\draw[draw=none,fill=green!50.1960784313725!black] (axis cs:10.8,0) rectangle (axis cs:11.2,10.32);
\draw[draw=none,fill=green!50.1960784313725!black] (axis cs:11.8,0) rectangle (axis cs:12.2,30.03);

\nextgroupplot[
tick align=outside,
tick pos=left,
x grid style={white!82.7450980392157!black},
xlabel={Features},
xmajorgrids,
xmin=0.18, xmax=13.82,
xtick style={color=black},
y grid style={white!82.7450980392157!black},
ymajorgrids,
ymin=0, ymax=0.753870768,
ytick style={color=black},
height=3cm, width=4cm
]
\draw[draw=none,fill=color0] (axis cs:0.8,0) rectangle (axis cs:1.2,0.01984089);
\draw[draw=none,fill=color0] (axis cs:1.8,0) rectangle (axis cs:2.2,0.03468785);
\draw[draw=none,fill=color0] (axis cs:2.8,0) rectangle (axis cs:3.2,0.0349181);
\draw[draw=none,fill=color0] (axis cs:3.8,0) rectangle (axis cs:4.2,0.01739891);
\draw[draw=none,fill=color0] (axis cs:4.8,0) rectangle (axis cs:5.2,0.0202103);
\draw[draw=none,fill=color0] (axis cs:5.8,0) rectangle (axis cs:6.2,0.02073888);
\draw[draw=none,fill=color0] (axis cs:6.8,0) rectangle (axis cs:7.2,0.01468042);
\draw[draw=none,fill=color0] (axis cs:7.8,0) rectangle (axis cs:8.2,0.02100945);
\draw[draw=none,fill=color0] (axis cs:8.8,0) rectangle (axis cs:9.2,0.03250469);
\draw[draw=none,fill=color0] (axis cs:9.8,0) rectangle (axis cs:10.2,0.01495992);
\draw[draw=none,fill=color0] (axis cs:10.8,0) rectangle (axis cs:11.2,0.02296499);
\draw[draw=none,fill=color0] (axis cs:11.8,0) rectangle (axis cs:12.2,0.02811335);
\draw[draw=none,fill=color0] (axis cs:12.8,0) rectangle (axis cs:13.2,0.71797216);

\nextgroupplot[
tick align=outside,
tick pos=left,
x grid style={white!82.7450980392157!black},
xlabel={Features},
xmajorgrids,
xmin=0.18, xmax=13.82,
xtick style={color=black},
y grid style={white!82.7450980392157!black},
ymajorgrids,
ymin=0, ymax=1.02421116,
ytick style={color=black},
height=3cm, width=4cm
]
\draw[draw=none,fill=red] (axis cs:0.8,0) rectangle (axis cs:1.2,0.0020309);
\draw[draw=none,fill=red] (axis cs:1.8,0) rectangle (axis cs:2.2,0.0021275);
\draw[draw=none,fill=red] (axis cs:2.8,0) rectangle (axis cs:3.2,0.00266701);
\draw[draw=none,fill=red] (axis cs:3.8,0) rectangle (axis cs:4.2,0.00227834);
\draw[draw=none,fill=red] (axis cs:4.8,0) rectangle (axis cs:5.2,0.00336496);
\draw[draw=none,fill=red] (axis cs:5.8,0) rectangle (axis cs:6.2,0.00206712);
\draw[draw=none,fill=red] (axis cs:6.8,0) rectangle (axis cs:7.2,0.00124582);
\draw[draw=none,fill=red] (axis cs:7.8,0) rectangle (axis cs:8.2,0.0020106);
\draw[draw=none,fill=red] (axis cs:8.8,0) rectangle (axis cs:9.2,0.00178229);
\draw[draw=none,fill=red] (axis cs:9.8,0) rectangle (axis cs:10.2,0.00129375);
\draw[draw=none,fill=red] (axis cs:10.8,0) rectangle (axis cs:11.2,0.00194976);
\draw[draw=none,fill=red] (axis cs:11.8,0) rectangle (axis cs:12.2,0.00174279);
\draw[draw=none,fill=red] (axis cs:12.8,0) rectangle (axis cs:13.2,0.9754392);

\nextgroupplot[
tick align=outside,
tick pos=left,
x grid style={white!82.7450980392157!black},
xlabel={Features},
xmajorgrids,
xmin=0.18, xmax=13.82,
xtick style={color=black},
y grid style={white!82.7450980392157!black},
ymajorgrids,
ymin=0, ymax=1.001864115,
ytick style={color=black},
height=3cm, width=4cm
]
\draw[draw=none,fill=blue] (axis cs:0.8,0) rectangle (axis cs:1.2,0.00373222);
\draw[draw=none,fill=blue] (axis cs:1.8,0) rectangle (axis cs:2.2,0.0028927);
\draw[draw=none,fill=blue] (axis cs:2.8,0) rectangle (axis cs:3.2,0.00394026);
\draw[draw=none,fill=blue] (axis cs:3.8,0) rectangle (axis cs:4.2,0.00453374);
\draw[draw=none,fill=blue] (axis cs:4.8,0) rectangle (axis cs:5.2,0.00500528);
\draw[draw=none,fill=blue] (axis cs:5.8,0) rectangle (axis cs:6.2,0.00337736);
\draw[draw=none,fill=blue] (axis cs:6.8,0) rectangle (axis cs:7.2,0.00435364);
\draw[draw=none,fill=blue] (axis cs:7.8,0) rectangle (axis cs:8.2,0.00325705);
\draw[draw=none,fill=blue] (axis cs:8.8,0) rectangle (axis cs:9.2,0.00365072);
\draw[draw=none,fill=blue] (axis cs:9.8,0) rectangle (axis cs:10.2,0.00413064);
\draw[draw=none,fill=blue] (axis cs:10.8,0) rectangle (axis cs:11.2,0.00343151);
\draw[draw=none,fill=blue] (axis cs:11.8,0) rectangle (axis cs:12.2,0.00353858);
\draw[draw=none,fill=blue] (axis cs:12.8,0) rectangle (axis cs:13.2,0.9541563);
\end{groupplot}

\end{tikzpicture}

\caption{Importance scores for (a) Perplexity Predictor, and (b)-(d) Latency predictor for 1080Ti, K80 GPUs and Xeon CPUs respectively. The features for (a) is the shape $S$, i.e., the dimensions across the 12 layers, while the latency predictor uses parameter count as a feature in addition to the shape $S$.}
\label{fig:importance_latency}
\end{figure*}
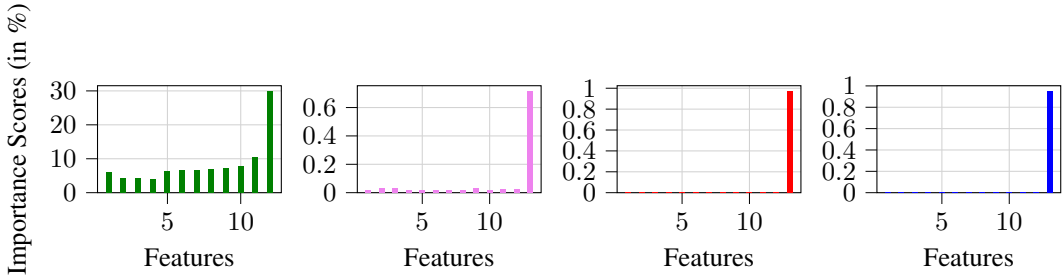

\noindent \textbf{Shape difference vs performance.}
To further study the effect of shape on performance, we test if the shape difference between random subnetworks and an optimal subnetwork (determined by evolutionary search) in the same parameter range, correlates with their differences in performance. The shape difference between two subnetworks with shapes $S_1$ and $S_2$ and their respective difference in performance (\gdirect) is characterised by their L2 norm of their differences:

\begin{equation}
\label{eq:}
    Diff =  \| S_1 - S_2 \|
\end{equation}

We choose points across different parameter ranges (50-100M) from the 10,000 random sampled subnetworks from section \Cref{sec:results} and compute their shape and performance differences with the optimal evolutionary-search model. The Spearman and Pearson correlation coefficient \cite{myers2004spearman, benesty2009pearson} across the shape and performance L2 norms are detailed in Table \Cref{tab:l2norm}. 
Clearly, we see a positive correlation between shapes and performance further reinstating the sensitivity of shape in determining optimal performance of a model. 

\begin{figure*}[!ht]
\centering

\begin{tikzpicture}

\definecolor{color0}{rgb}{0,0,1}
\definecolor{color1}{rgb}{0,1,0}

\begin{groupplot}[group style={group size=3 by 1}]
\nextgroupplot[
legend cell align={left},
legend style={fill opacity=0.8, draw opacity=1, text opacity=1, draw=white!80!black},
tick align=outside,
tick pos=left,
x grid style={white!82.7450980392157!black},
xlabel={Latency (in s)},
xmajorgrids,
xmin=0.219, xmax=0.461,
xtick style={color=black},
y grid style={white!82.7450980392157!black},
ylabel={Perplexity},
ymajorgrids,
ymin=5.3475, ymax=12.0025,
ytick style={color=black},
height=4cm, width=6cm
]
\addplot [ultra thick, color0, mark=*, mark size=1, mark options={solid}]
table {%
0.240000009536743 11.6999998092651
0.25 10.9700002670288
0.289999961853027 8.59000015258789
0.330000042915344 6.98000001907349
0.350000023841858 6.61999988555908
0.389999985694885 6.17000007629395
0.430000066757202 5.82999992370605
0.450000047683716 5.65000009536743
};
\addplot [ultra thick, red, mark=*, mark size=1, mark options={solid}]
table {%
0.240000009536743 10.9700002670288
0.296000003814697 7.8899998664856
0.330000042915344 6.78999996185303
0.360000014305115 6.26000022888184
0.419999957084656 5.73999977111816
};
\addplot [ultra thick, color1, mark=*, mark size=1, mark options={solid}]
table {%
0.409999966621399 5.69999980926514
0.409999966621399 5.73999977111816
0.350000023841858 6.48000001907349
0.319999933242798 7.13000011444092
0.309999942779541 7.32999992370605
0.299999952316284 7.8600001335144
0.259999990463257 9.51000022888184
0.25 10.9700002670288
0.230000019073486 11.4399995803833
};

\nextgroupplot[
legend cell align={left},
legend style={fill opacity=0.8, draw opacity=1, text opacity=1, draw=white!80!black},
scaled y ticks=manual:{}{\pgfmathparse{#1}},
tick align=outside,
tick pos=left,
x grid style={white!82.7450980392157!black},
xlabel={Latency (in s)},
xmajorgrids,
xmin=0.122, xmax=0.298,
xtick style={color=black},
y grid style={white!82.7450980392157!black},
ymajorgrids,
ymin=5.3475, ymax=12.0025,
ytick style={color=black},
yticklabels={},
height=4cm, width=6cm
]
\addplot [ultra thick, color0, mark=*, mark size=1, mark options={solid}]
table {%
0.139999985694885 10.9700002670288
0.169999957084656 8.59000015258789
0.200000047683716 6.98000001907349
0.220000028610229 6.61999988555908
0.240000009536743 6.17000007629395
0.269999980926514 5.82999992370605
0.289999961853027 5.65000009536743
};
\addplot [ultra thick, red, mark=*, mark size=1, mark options={solid}]
table {%
0.157000064849854 9.38000011444092
0.159999966621399 8.86999988555908
0.169999957084656 7.8899998664856
0.230000019073486 6.42000007629395
};
\addplot [ultra thick, color1, mark=*, mark size=1, mark options={solid}]
table {%
0.269999980926514 5.69999980926514
0.259999990463257 5.73999977111816
0.220000028610229 6.48000001907349
0.190000057220459 7.13000011444092
0.190000057220459 7.32999992370605
0.180000066757202 7.8600001335144
0.149999976158142 9.51000022888184
0.129999995231628 10.9700002670288
0.129999995231628 11.4399995803833
};

\nextgroupplot[
legend cell align={left},
legend style={fill opacity=0.8, draw opacity=1, text opacity=1, draw=white!80!black},
scaled y ticks=manual:{}{\pgfmathparse{#1}},
tick align=outside,
tick pos=left,
x grid style={white!82.7450980392157!black},
xlabel={Latency (in s)},
xmajorgrids,
xmin=0.82145, xmax=1.86755,
xtick style={color=black},
y grid style={white!82.7450980392157!black},
ymajorgrids,
ymin=5.3475, ymax=12.0025,
ytick style={color=black},
yticklabels={},
height=4cm, width=6cm
]
\addplot [ultra thick, color0, mark=*, mark size=1, mark options={solid}]
table {%
0.86899995803833 11.6999998092651
0.934999942779541 10.9700002670288
1.07000005245209 8.59000015258789
1.25399994850159 6.98000001907349
1.3400000333786 6.61999988555908
1.53600001335144 6.17000007629395
1.68700003623962 5.82999992370605
1.76999998092651 5.65000009536743
};
\addlegendentry{Parameters Constrained}
\addplot [ultra thick, red, mark=*, mark size=1, mark options={solid}]
table {%
1.02999997138977 9.27000045776367
1.12999999523163 8.11999988555908
1.22000002861023 7.36999988555908
1.36000001430511 6.5
1.4099999666214 6.23999977111816
};
\addlegendentry{Latency Constrained}
\addplot [ultra thick, color1, mark=*, mark size=1, mark options={solid}]
table {%
1.82000005245209 5.69999980926514
1.75999999046326 5.73999977111816
1.49000000953674 6.48000001907349
1.33000004291534 7.13000011444092
1.29999995231628 7.32999992370605
1.24000000953674 7.8600001335144
1.08000004291534 9.51000022888184
0.970000028610229 10.9700002670288
0.889999985694885 11.4399995803833
};
\addlegendentry{Heuristically shaped}
\end{groupplot}

\end{tikzpicture}

\caption{Perplexity vs Latency for optimal models searched using EA with parameter and latency constrained and for heuristically shaped models across: (a) 1080Ti GPU, (b) Xeon CPU, and (c) K80 GPU}
\label{fig:lat_pareto}
\end{figure*}
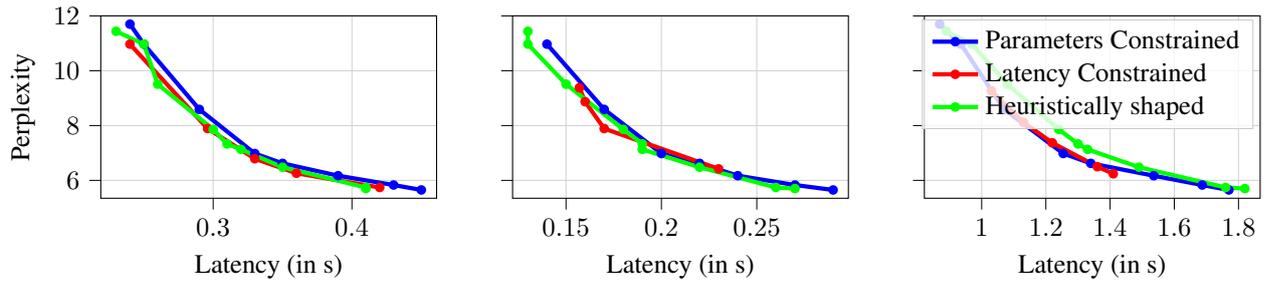
\input{Images/predictor_performances}

\begin{figure}[!ht]
    \centering
\begin{tikzpicture}

\definecolor{color0}{rgb}{0.5,0,1}
\definecolor{color1}{rgb}{0.280392156862745,0.338158274815817,0.985162233467507}
\definecolor{color2}{rgb}{0.0607843137254902,0.636474236147141,0.941089252501372}
\definecolor{color3}{rgb}{0.166666666666667,0.866025403784439,0.866025403784439}
\definecolor{color4}{rgb}{0.386274509803922,0.984086337302604,0.767362681448697}
\definecolor{color5}{rgb}{0.613725490196078,0.984086337302604,0.641213314833578}
\definecolor{color6}{rgb}{0.833333333333333,0.866025403784439,0.5}
\definecolor{color7}{rgb}{1,0.636474236147141,0.338158274815817}
\definecolor{color8}{rgb}{1,0.338158274815818,0.17162567916636}

\begin{axis}[
legend cell align={left},
legend columns=3,
legend style={
  fill opacity=1,
  draw opacity=1,
  text opacity=1,
  at={(0.04,1)},
  anchor=north west,
  draw=white!80!black,
  nodes={scale=0.55, transform shape}
},
tick align=outside,
tick pos=left,
x grid style={white!69.0196078431373!black},
xlabel={Layers},
xmin=0.45, xmax=12.55,
xtick style={color=black},
y grid style={white!69.0196078431373!black},
ylabel={Hidden Sizes},
ymin=87.6, ymax=800.4,
ytick style={color=black},
height=4cm, width=8cm
]
\addplot [thick, color0, mark=*, mark size=1, mark options={solid}]
table {%
1 600
2 360
3 360
4 360
5 600
6 600
7 600
8 600
9 768
10 768
11 768
12 768
};
\addlegendentry{91M}
\addplot [thick, color1, mark=*, mark size=1, mark options={solid}]
table {%
1 540
2 360
3 360
4 360
5 540
6 540
7 540
8 540
9 768
10 768
11 768
12 768
};
\addlegendentry{87M}
\addplot [thick, color2, mark=*, mark size=1, mark options={solid}]
table {%
1 480
2 240
3 240
4 240
5 480
6 480
7 480
8 480
9 540
10 540
11 600
12 600
};
\addlegendentry{70M}
\addplot [thick, color3, mark=*, mark size=1, mark options={solid}]
table {%
1 360
2 240
3 240
4 240
5 360
6 360
7 360
8 360
9 480
10 480
11 540
12 540
};
\addlegendentry{61M}
\addplot [thick, color4, mark=*, mark size=1, mark options={solid}]
table {%
1 360
2 240
3 240
4 240
5 360
6 360
7 360
8 360
9 360
10 360
11 540
12 540
};
\addlegendentry{59M}
\addplot [thick, color5, mark=*, mark size=1, mark options={solid}]
table {%
1 360
2 120
3 120
4 120
5 360
6 360
7 360
8 360
9 360
10 360
11 540
12 540
};
\addlegendentry{56M}
\addplot [thick, color6, mark=*, mark size=1, mark options={solid}]
table {%
1 240
2 120
3 120
4 240
5 240
6 240
7 240
8 240
9 360
10 360
11 360
12 360
};
\addlegendentry{47M}
\addplot [thick, color7, mark=*, mark size=1, mark options={solid}]
table {%
1 240
2 120
3 120
4 120
5 240
6 240
7 240
8 240
9 240
10 240
11 240
12 240
};
\addlegendentry{42M}
\addplot [thick, color8, mark=*, mark size=1, mark options={solid}]
table {%
1 120
2 120
3 120
4 120
5 120
6 120
7 120
8 120
9 240
10 240
11 240
12 240
};
\addlegendentry{37M}
\end{axis}

\end{tikzpicture}
\caption{Heuristically shaped models have a cigar-like shape}
\label{fig:heuristic_shapes}
\vspace{-2em}
\end{figure}
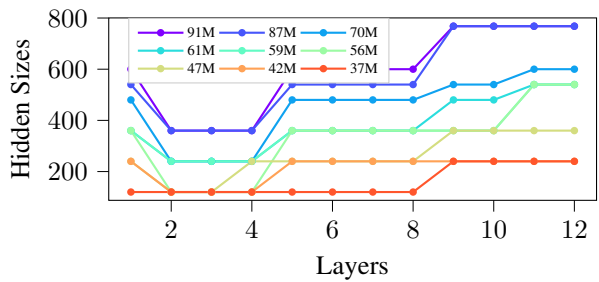

\noindent \textbf{Feature importance from optimal sub-networks.}

From the analysis of sub-networks searched by EA and the templated shapes, we find that accurate networks have more parameters in later layers.
We analyse this using the perplexity predictor trained to estimate \gpartial~given the shape.
For this predictor, we compute the feature importance (see \Cref{fig:importance_latency}(a)) of each layer's shape and find it to be highest for the last few layers and the first layer.
Based on these observations, we derive a set of \textit{heuristics} of what is a good shape: (a) a large dimension in the last layer, (b) moderately large dimension in the first layer, (c) low dimensions in early middle layers (2-5), and (d) moderate dimensions in later middle layers (6-11). 
We characterize this as a \textit{cigar-like shape}.


%

\noindent \textbf{Heuristically shaped models.}
Based on the above heuristics, we hand-shape sub-networks with the following algorithm: 
(a) construct a reference model $T_{S^*}$ following the heuristics at a given parameter range (60-65M),
(b) for a target parameter count, scale the shape $S_i$ of every layer linearly,
(c) for early middle layers, round down the scaled $S_i$ and for remaining layers round up $S_i$ to the nearest configuration in $D$.
Based on this algorithm, we identify sub-networks across the parameter count with cigar-like shapes as shown in \Cref{fig:heuristic_shapes}.
We evaluate these hand-crafted sub-networks on perplexity and find that they are competitive and even outperform sub-networks searched with EAs (see \Cref{fig:pareto_heuristic}). We also pre-train and evaluate one of the heuristic models with a parameter count of 61M on the Glue tasks (see Table \ref{tab:supershaper-best}). We observe that, similar to our evolutionary-search subnetwork (63M), the heuristic model outperforms prominent hand-crafted or compressed networks and only lags behind NAS-BERT and DynaBERT.
This strongly demonstrates the generalization of the derived heuristics across model size.
To the best or our knowledge, this is the first such generalization demonstrated for NAS.

\subsection{Device-specific efficient models.}

We now discuss searching for sub-networks based on latency on a device.
We train latency predictors for 3 devices - 1080Ti, K80, and a Xeon PC with high $R^2$ values of $0.993$, $0.988$ and $0.892$ respectively as shown by the fitness curves in \Cref{fig:predictor_performances}. 
The feature importance of these predictors strongly favours total parameters (96.5\%) and only very weakly depends on layer dimensions as shown in \Cref{fig:importance_latency}(b),(c) and (d).
This is a crucial insight: the shape of the network for a given parameter count is a free variable that can be optimized for accuracy. 
Thus for deployment on a device, we need to identify the right parameter count that meets the latency constraint while the shape can be identified with EA or the heuristics we have laid out.

We run EA under two settings - parameter constraints and latency constraints for all  devices. 
We also evaluate the hand-crafted models.
The latency and perplexity of these models are shown in \Cref{fig:lat_pareto}.
As can be seen, all three techniques result in similar performance.
This corroborates that latency is insensitive to shape and that the heuristics identify competitive networks. 



In summary, we showed that SuperShaper effectively generalizes training across sub-networks, and finds competitive networks at various sizes.
This training on language models enables generalization across tasks.
Further we derived a set of simple rules to shape a network which is competitive with EA search, thereby easily generalizing the search across model size.
And finally we established that latency on devices is insensitive to shapes and thus EA search on parameter count or hand-crafted networks generalize across devices.
Thus, with a simple and effective super-pretraining procedure we identify sub-networks that generalize across tasks, model sizes, and devices.

\section{Conclusions and Future Work}
\label{sec:conclusions}
To address the problem of deploying NLU models across a range of devices, we propose SuperShaper a NAS technique to pretrain language models by shaping Transformer layers.
SuperShaper identifies networks that outperform state-of-the-art model compression techniques on GLUE benchmarks.
We discovered that cigar-like shapes of networks generalize across parameter counts and device latency is insensitive to shape.
Consequently, we demonstrate that NAS can be performed with radically simple design space and implementation, while deriving generalization across tasks, model sizes, and devices.
This work can be extended (a) to other tasks such as NLG, and (b) to generate smaller models in combination with other compression techniques.

\section{Acknowledgements}
We thank  AI4Bharat and Robert Bosch Centre for Data Science and Artificial Intelligence, IIT Madras for their help with compute resources for this project. We thank wandb \cite{wandb} for providing a very useful experiment tracking tool that helped accelerate our progress. We thank Vamsi Krishna for useful discussions in the early stages of this project.


%
%
%

\bibliography{cite}
\bibliographystyle{acl_natbib}

\end{document}